\documentclass[10pt,letterpaper]{article}

\usepackage{cogsci}
\usepackage{pslatex}
\usepackage{apacite}
\usepackage{natbib}
\usepackage{graphicx}
\usepackage{verbatim}
\usepackage{listings}
\usepackage{array}
\usepackage{booktabs}
\usepackage[hidelinks]{hyperref}
\usepackage{cleveref}
\usepackage{microtype}
\usepackage{amssymb} 
\usepackage{placeins}
\usepackage{tabularx}
\usepackage{tcolorbox}

\newtcolorbox{examplebox}[2][]{
    colback=gray!10,
    colframe=gray!80,
    coltitle=white,
    fontupper=\footnotesize,
    fonttitle=\bfseries,
    title=#2,                      
    #1
}

\newcommand{\up}{\textcolor{green!60!black}{$\blacktriangle$}}
\newcommand{\down}{\textcolor{red}{$\blacktriangledown$}}

\usepackage{xcolor} 

\lstdefinestyle{PromptStyle}{
    basicstyle=\ttfamily\small\color{black},
    language={},
    columns=fullflexible,
    breaklines=true,
    breakindent=0pt,
    showstringspaces=false,
    tabsize=2,
    frame=none,
    backgroundcolor=\color{white},
    numbers=none,
}

\title{Greedy or not, here I come: Language production under \\ vocabulary constraints in humans and resource-rational models}

\author{
  {\Large \textbf{Thomas Hikaru Clark}$^{1}$, \textbf{Sihan Chen}$^{1}$,
  \textbf{Laura Nicolae}$^{2}$}
  \\
  {\normalsize $^{1}$Department of Brain and Cognitive Sciences, MIT} \hspace{5mm}
  {\normalsize $^{2}$Department of Economics, Harvard University}
  \\
  {\normalsize \texttt{\{thclark, sihanc\}@mit.edu, lauranicolae@g.harvard.edu}}
}

\begin{document}

\maketitle

\begin{abstract}

Communicating using only a limited vocabulary is a common but challenging cognitive phenomenon, requiring an ideal communicator to plan carefully to optimize for intelligibility while circumventing a constrained lexicon. 
In this work, we investigate how humans respond to a broad array of questions under variable vocabulary limitations, consisting of only 250 highly frequent words at the most restrictive.
We provide theoretically motivated comparisons to greedy and globally optimal sampling algorithms using Sequential Monte Carlo inference with large language models. 
Humans generally resemble greedy sampling more than globally optimal sampling, though more skilled humans are more likely to backtrack and revise -- a non-greedy behavior.
An observed human pattern of leaning on semantically light words in high-constraint settings falls out of both greedy and globally optimal sampling.
We discuss the results and their broader implications for resource-rational cognition, psycholinguistics, L2 communication, and language impairments. 

\end{abstract}

\section{Introduction}

Humans use language to communicate  \citep{fedorenkoLanguagePrimarilyTool2024}.
Using insights from information theory \citep{shannonMathematicalTheoryCommunication1948}, researchers model language as communication code between a speaker and a listener: the speaker converts their meaning to utterances, and the listener receives the utterances and recovers the meaning \citep[see][for a review]{gibsonHowEfficiencyShapes2019}.
The production process is incremental: instead of starting with a full utterance in mind, a person typically plans what to say as they speak or write \citep[e.g.][]{bockLanguageProductionGrammatical1994, ferreiraHowIncrementalLanguage2002, pickeringIntegratedTheoryLanguage2013}.
At the computational level, \citet{futrellInformationtheoreticPrinciplesIncremental2023} modeled language production as an action planning problem in which a speaker's incremental production is influenced both by a pressure to say contextually predictable things and a pressure to say things that further the communicative goal. 
In this work, we specifically investigate how constraints on the available set of vocabulary words affect incremental language production.

Communication under a vocabulary constraint is a common cognitive phenomenon.
Many societies contain linguistic minorities with limited knowledge of the majority language, which affects their ability to access basic services or participate in the economy \citep{bleakleyLanguageSkillsEarnings2004, groggerWagePenaltyRegional2020}.
Compared to native speakers, learners of a language (L2 learners) tend to use fewer and more frequent words \citep{lauferDevelopmentL2Lexis1991}.
When L2 learners do not know a specific word, they may employ various strategies to get a meaning across, such as describing key properties of the concept \citep{poulisseTheoreticalAccountLexical2011} or using ``all-purpose'' or semantically light words as substitutes \citep{dornyeiCommunicationStrategiesSecond1997}.
This phenomenon suggests that there may be diminishing marginal returns to learning new vocabulary words above a certain point, since a subset of the vocabulary is typically sufficient for communication.

Another common case of vocabulary-constrained communication is when an expert explains a complex phenomenon to a layperson.
Past work has considered what makes for a good explanation \citep[e.g.][]{brewerExplanationScientistsChildren2000, cruzHowLaypeopleEvaluate2025, chandraCooperativeExplanationRational2024, mccarthyRightWayExplain2023}.
For example, \citet{sulikExplanationsWild2023} suggests that a good explanation contains functional (i.e., what is something for?) or mechanistic (i.e., how does something work?) information.
Besides the type of information, the words used also matter. For example, laypersons find explanations with jargon (terms only understood by a specific group of experts) harder to understand \citep{bullockJargonBarrierEffective2019, cruzHowLaypeopleEvaluate2025, keuleersWordKnowledgeCrowd2015}.
It remains unclear how exactly the usage of specific words changes in response to vocabulary constraints, and what kind of algorithmic-level \citep{marrVisionComputationalInvestigation1982} model of language generation captures the properties of language generated under vocabulary constraints. 

In this work, we define a large space of communicative goals and evaluate the properties of English responses generated by humans under varying vocabulary constraints, comprising only the most common 250 English words at the most restrictive.
Recently-developed techniques allow us to sample from a language model subject to custom constraints \citep{lipkinFastControlledGeneration2025, loulaSyntacticSemanticControl2025}, and provide a new opportunity to test algorithmic-level hypotheses about human cognition. 
In particular, we can manipulate the \textit{greediness} of constrained generation, where a purely greedy algorithm has a bias for locally high-probability continuations, while approximately globally-optimal inference algorithms like particle-based Sequential Monte Carlo (SMC) avoid these biases and are thus more consistent with planning ahead during language production. 
We investigate whether humans perform constrained language generation in a way that aligns more closely with greedy generation or globally-optimal inference.

To foreshadow our results: for the humans in our study, performance as a function of vocabulary size more closely resembled greedy sampling than SMC-based constrained generation, which handled very small vocabularies better than humans or greedy sampling. 
Despite this, we observe that top-scoring humans revise valid string prefixes (a non-greedy behavior) significantly more than low-scoring humans. This suggests that human approaches to constrained generation are heterogeneous: some plan ahead or revise their answers while others prioritize simple, greedy, and imperfect responses. 
We also observe interpretable patterns in the frequency of word use under different vocabulary constraints: semantically ``light'' words, like \textit{do, thing,} and \textit{people}, are used disproportionately frequently under strict vocabulary constraints, even relative to other allowed words. This pattern is reproduced by constrained generation from language models. 
We conclude by discussing the implications of these findings for both psycholinguistics and for broader areas of impact. 

\section{Methods}

\subsection{Constrained vocabulary definition}

We define a constrained vocabulary of size $N$, consisting of the $N$ most frequent words in English, according to the word list provided by the \texttt{wordfreq} Python package \citep{speerRspeerWordfreqV302022a}. For each word in the list, the set of words sharing the same base form (lemma) was also included in the list, using the \texttt{lemminflect} Python package, available within \texttt{spaCy} \citep{honnibalSpaCyIndustrialstrengthNatural2020}. For example, if any of the word forms \textit{drink, drank, drunk, drinks,} or  \textit{drinking} were present within the top $N$ words, then all word forms in this set were included. We define seven vocabulary sets, starting at 250 words and doubling up to 16,000 words. 
For reference, it is estimated that the average native speaker of American English is familiar with 42,000 word lemmas from 11,000 word families \citep{brysbaertHowManyWords2016}.

\subsection{Questions dataset}

To simulate a variety of real-world situations with diverse communicative goals, we assemble a dataset of 192 questions divided into the following four categories of 48 questions each: Why, How, ExplainSimple, and RedditELI5 (Explain Like I'm 5).
The Why questions are sourced from a study by \citet{sulikExplanationsWild2023}, who investigated which features make for good explanations. 
The How and ExplainSimple datasets were created from scratch using fixed sentence patterns (``How is/are/do/does/can/would [BLANK]?'' and ``Explain [BLANK] in simple terms''), with the goal of covering a wide range of communicative topics, including everyday life, sports, and science.  
The RedditELI5 questions were sourced from the all-time most popular questions on the Reddit forum ``Explain Like I'm 5'', lightly edited for length and clarity. 

\subsection{Human behavioral experiment}

We created an online behavioral experiment in which participants respond to questions from the above dataset, using an interface which allows them to type only words within a specified vocabulary. While not a fully natural task, the interface simulates a speaker encountering vocabulary limitations, forcing on-line adaptation to the constraint via circumlocution or other strategies. 
Participants could only type or delete (selecting, replacing, and inserting text was blocked). 
We recruited 144 English speakers from Prolific, who were each paid \$6. The study was conducted under an approved IRB protocol at the authors' institution. Each participant answered 16 questions, beginning with 4 questions with no vocabulary constraints, then 4 questions each for vocabulary sizes of 4000, 1000, and 250. Each question in the dataset was answered by three unique individuals at each vocabulary size. 
Participants were prompted to move to the next question after 90 seconds without a submission. 
We record participants' final responses as well as all intermediate keystrokes.

\subsection{Constrained LLM generation with \textsc{Awrs}}

Adaptive Weighted Rejection Sampling \citep{lipkinFastControlledGeneration2025} is a Sequential Monte Carlo method for sampling incrementally from a language model, subject to user-defined constraints (implemented as binary functions that can be evaluated on partially generated strings). 
It does this by maintaining multiple hypotheses in parallel, represented by weighted \textit{particles}. 
We use \textsc{Awrs} to generate responses to prompts, using a custom potential in the \texttt{GenLM-control} library to enforce a hard constraint on the set of words in the vocabulary. 
Constrained generation was performed using the \texttt{Llama-3.2-1B-Instruct} model, with either 16 or 32 particles for SMC inference (denoted \textsc{Awrs-16} and \textsc{Awrs-32}, respectively). By setting the number of particles to 1, the algorithm is equivalent to local greedy sampling, where only one hypothesis is maintained at each step of generation.
A prompt with instructions and two few-shot examples was provided. 
Example responses from both humans and models, under both a permissive and a restrictive constraint, are shown in \Cref{tab:example_responses_8850}.

\subsection{Analyses}

\paragraph{Automated evaluation of response quality}

To create an estimate of the quality of responses to each question (from both humans and models), we use a prompted LLM (\texttt{Llama-3.1-8B-Instruct}) to assign scores on a 7-point Likert scale, along with accompanying justifications, under an LLM-as-judge paradigm \citep{guSurveyLLMasaJudge2025, zhengJudgingLLMasaJudgeMTBench2023}. 
For each question, we aim to approximate the value $\mathbb{E}[f(X)] = \sum_x f(X) p_X(x) dx$ where $X$ denotes a random variable sampled from the distribution over constrained responses to a question, $f(\cdot)$ denotes the scoring function, and $p_X$ denotes the probability density function of X, approximated by the normalized SMC weights. For computational efficiency, we discard samples from SMC with weights below a chosen threshold of 0.01, which contribute minimally to the sum. 
We use the same evaluation pipeline on human responses, treating each response as a single sample (all equally weighted). 
The automated evaluator is an imperfect proxy for response quality that was employed to address the large number of both model and human responses that required annotation, for which it was not practical to collect comprehensive human judgments. 
While the absolute score assigned to human- vs. model-generated utterances may be biased (e.g. if LLM judges prefer LLM-generated responses), we are primarily concerned with \textit{changes} to the evaluation score as a function of vocabulary size.
We conducted a norming study to validate the LLM-as-judge pipeline: one third of questions (64 out of 192) were randomly sampled, and for each of four different vocabulary sizes, one response from humans, the greedy model, and the \textsc{Awrs-32} model was randomly chosen and graded by N=24 human participants on Prolific, who each saw 32 question-answer pairs (users who participated in the constrained generation experiment were excluded). Human graders were given the same instructions provided to the LLM-as-judge, rating the response quality on a 7-point Likert scale. For the subset of questions in the norming study, automated LLM-generated ratings and human ratings had a Spearman $\rho$ of 0.60, suggesting that the automated pipeline captures a substantial amount of variance in human ratings. 

We predict that the average response score will decrease monotonically as the allowed vocabulary is reduced, and we evaluate whether scores for human responses decline similarly to the greedy or SMC-based model responses as vocabulary size is restricted. 

\paragraph{Shifts in word frequency}

We compute the frequency of each word in the generated output as a function of vocabulary size. 
Restricting the vocabulary necessarily removes low-frequency words from the output distribution, but does not \textit{directly} change the relative frequencies of the remaining words. For example, all remaining words might be used slightly more frequently to account for the words that were removed from the vocabulary, but it is not obvious that there would be any change in the usage rank of the remaining words. 
However, we hypothesize that tightening the vocabulary constraint will systematically change the frequency distribution of used words, even among those that are allowed by the constraint. In particular, words which have a high degree of \textit{substitutability} with low-frequency words will increase in usage rank relative to words with a low degree of substitutability  \citep{varianIntermediateMicroeconomicsModern2010}. For example, in settings where low-frequency words are not available, we may expect the word ``thing'' to be substituted instead \citep{hasselgrenLexicalTeddyBears1994, kleinBasicVarietyCouldnt1997}. 
Meanwhile, function words, including prepositions, conjunctions, and determiners, may not see a large change in usage. 

\paragraph{Revision as an index of greediness vs. planning}

\begin{figure}[hbt]
\centering
  \includegraphics[width=0.95\columnwidth]{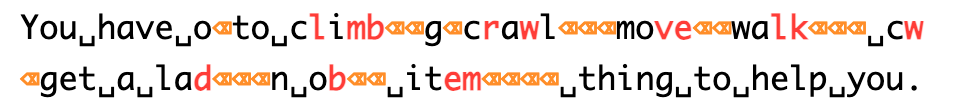}
  \caption {Example keystroke sequence under a 250-word vocabulary, in response to ``How do you rescue a cat from a tree?''. Red characters denote disallowed keystrokes; the backspace symbol is displayed in orange. The final response was ``\texttt{You have to get a thing to help you.}''.}
  \label{fig:example-keystrokes}
\end{figure}

Under strictly greedy language generation, once a word is output, it can no longer be revised. In language modeling terms, this feature of greedy sampling can lead to so-called ``dead ends,'' where a sequence of high-probability words leads to a context in which the only high-probability continuation is outside of the vocabulary \citep[see][for a discussion]{lewSequentialMonteCarlo2023}. For example, given the sequence ``It was a blessing in...'', Google N-grams estimates a 90\% probability that the next word is ``disguise''.
If this word is outside the vocabulary, a greedy sampler would be forced to choose a different, low-probability continuation, which could rapidly lead to an incoherent sentence being generated.
A rational constrained language generator could avoid this problem by either planning ahead (avoiding going down the dead end in the first place) or by revising (backtracking or deleting words once a dead end is detected). 
It is difficult to directly measure whether a person is planning ahead. However, given the fine-grained typing data from our online task (\Cref{fig:example-keystrokes}), it is possible to see whether participants ever delete previously written words. For a word to be counted as deleted, it must have first been completed (e.g., by a whitespace or punctuation character) and then erased by repeated presses of the backspace key. We compute the average number of word deletions per response to quantify revisions.

\section{Results}

\begin{table}[hbt]
\centering
\caption{Example responses at different vocabulary sizes.}
\vskip 0.12in

\footnotesize
\begin{tabularx}{\columnwidth}{p{2.0cm}>{\raggedright\arraybackslash}X}
\multicolumn{2}{p{\columnwidth}}{\textbf{Question:} Explain machine learning in simple terms.} \\
\toprule
Source (Vocab) & Response \\
\midrule
Human (4000) & Machine Learning is when a Machine is taught how to do specific tasks and assist with issues. \\
\midrule
Greedy (4000) & Machine learning is a way computers can learn from data without being told what to do. It's like teaching a child to recognize objects by showing them a lot of pictures. \\
\midrule
\textsc{Awrs-32} (4000) & Machine learning is a way computers can learn from data and get better at doing things on their own. It's like having a super smart assistant that can recognize patterns and make decisions automatically. \\
\midrule
Human (250) & Over and over it knows. \\
\midrule
Greedy (250) & It's a way that a big, very old, very good thinking game helps make things that are like us, like you, not that way. It uses big groups of people giving lots of old, old things into the game ...
\\
\midrule
\textsc{Awrs-32} (250) & It's a way of making things good at doing things on their own. You give it some old things it's good at, and it gets better over time. \\
\bottomrule
\end{tabularx}
\label{tab:example_responses_8850}
\end{table}

\subsection{Effects of vocabulary size reveal similarity between humans and greedy sampling}

\begin{figure}[htb]
\centering
  \includegraphics[width=\columnwidth]{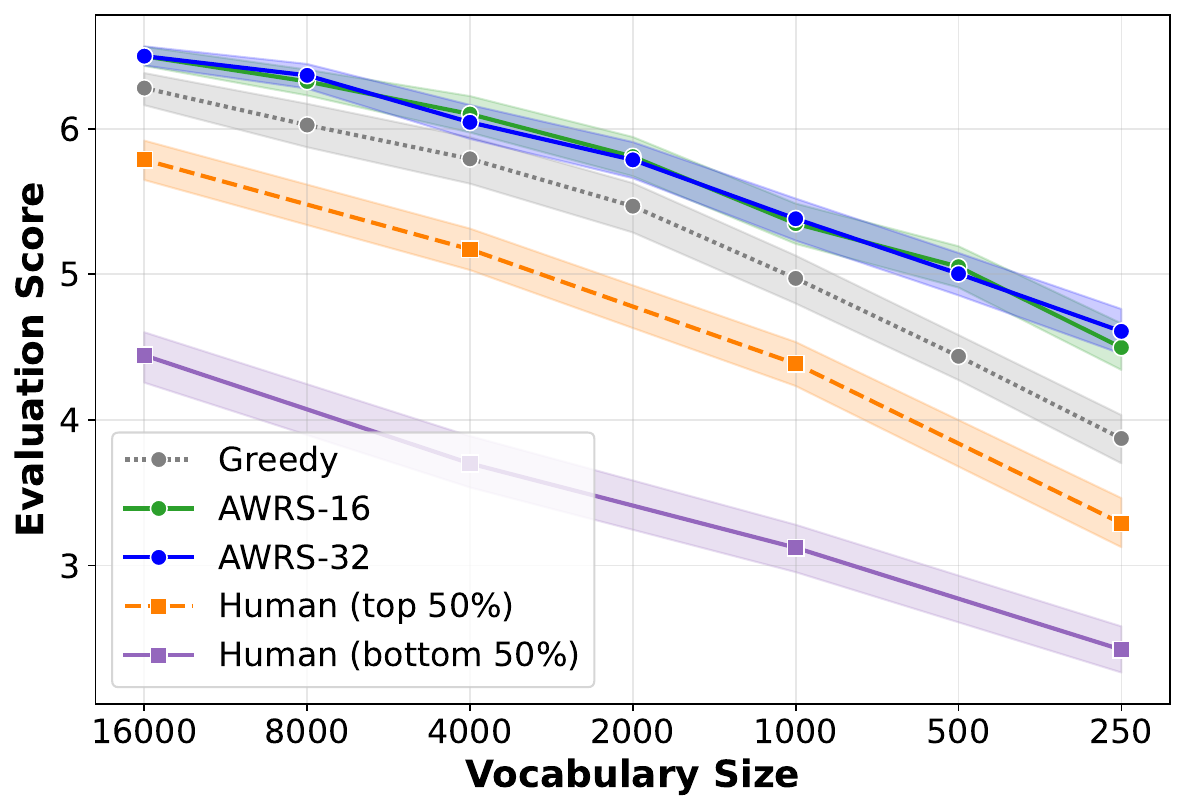}
  \caption {Mean weighted score (automated evaluation on a 1-7 Likert scale) as a function of allowed vocabulary size. The horizontal axis is logarithmic. Shaded bands denote 95\% confidence intervals over the scores for all 192 questions. The label `16000' is mapped to the unconstrained condition for humans.}
  \label{fig:vocab-vs-score}
\end{figure}

\Cref{fig:vocab-vs-score} shows the average evaluation scores of responses generated by humans and the three algorithms (greedy, \textsc{Awrs-16}, and \textsc{Awrs-32}) at each vocabulary size. 
Exclusion criteria for participants were not pre-defined, but we perform a post-hoc split of participants into the top 50\% and bottom 50\% by average response score to separate out low-effort responses.
In absolute terms, human responses scored lower, on average, than all models at each vocabulary size. This may be influenced by the fact that human responses were generally shorter and less detailed than model responses. All three algorithms and the human-generated responses exhibit diminishing marginal returns: evaluation scores increase by nearly-constant increments each time the vocabulary size is doubled, suggesting that the average marginal utility of adding a new word to the vocabulary declines as the vocabulary size expands. 

Tightening the vocabulary constraint produced similar decreases in the scores of the human responses and the greedy algorithm's responses at all vocabulary sizes. Meanwhile, the \textsc{Awrs} algorithm displays greater robustness to vocabulary constraints compared to humans and the greedy algorithm: the evaluation scores of \textsc{Awrs}-generated responses declined more slowly than those of humans and the greedy algorithm, with generally larger gaps in performance for smaller vocabularies. On the 250-word vocabulary, the \textsc{Awrs} algorithm achieves half a point better (on a 7-point scale) than would be expected if its performance degraded at the same rate as human participants and greedy sampling.

\subsection{A shift towards semantically light words in constrained generation}

\begin{figure}[!htb]
\centering
  \includegraphics[width=0.95\columnwidth]{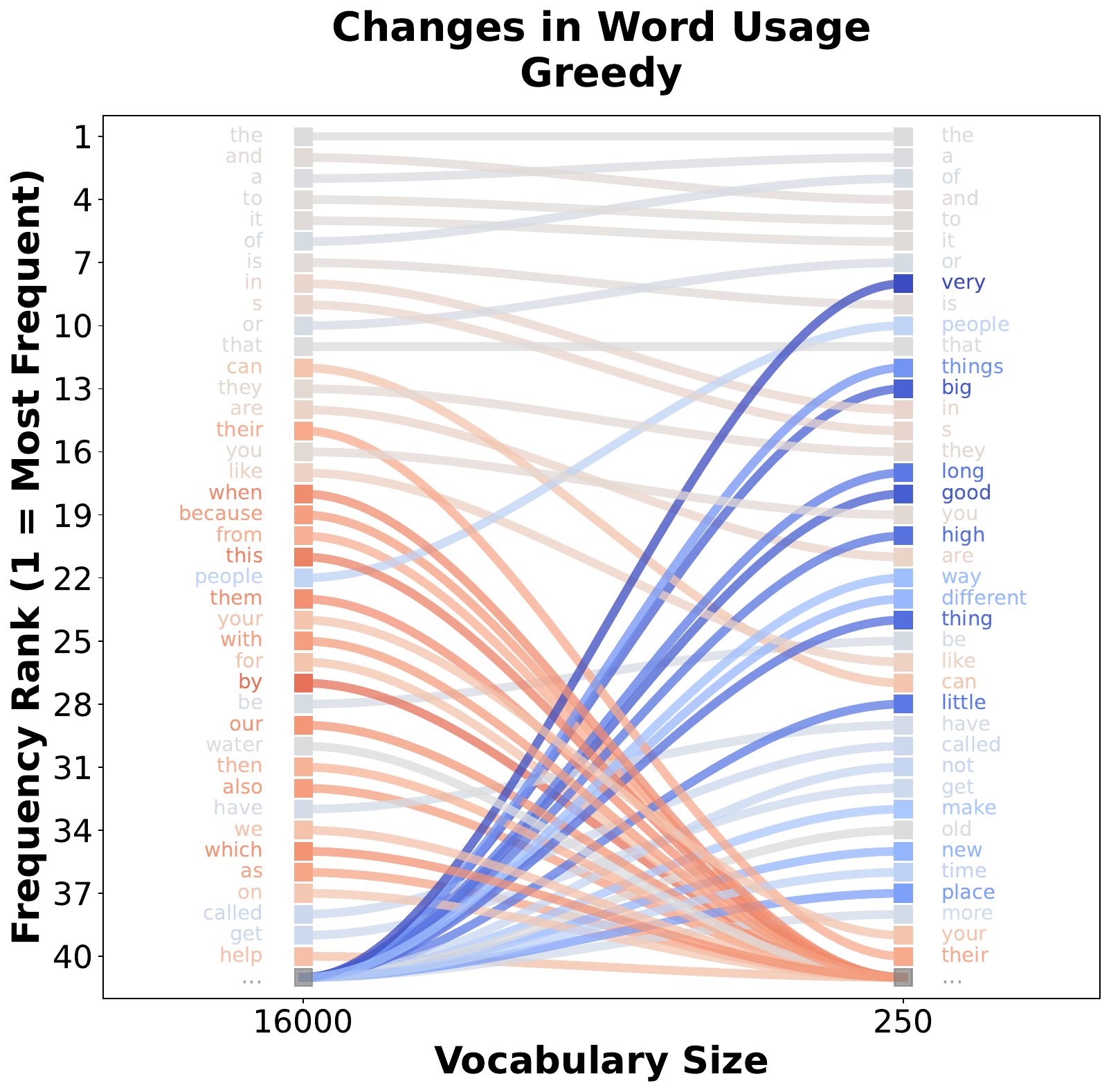}
  
  \includegraphics[width=0.95\columnwidth]{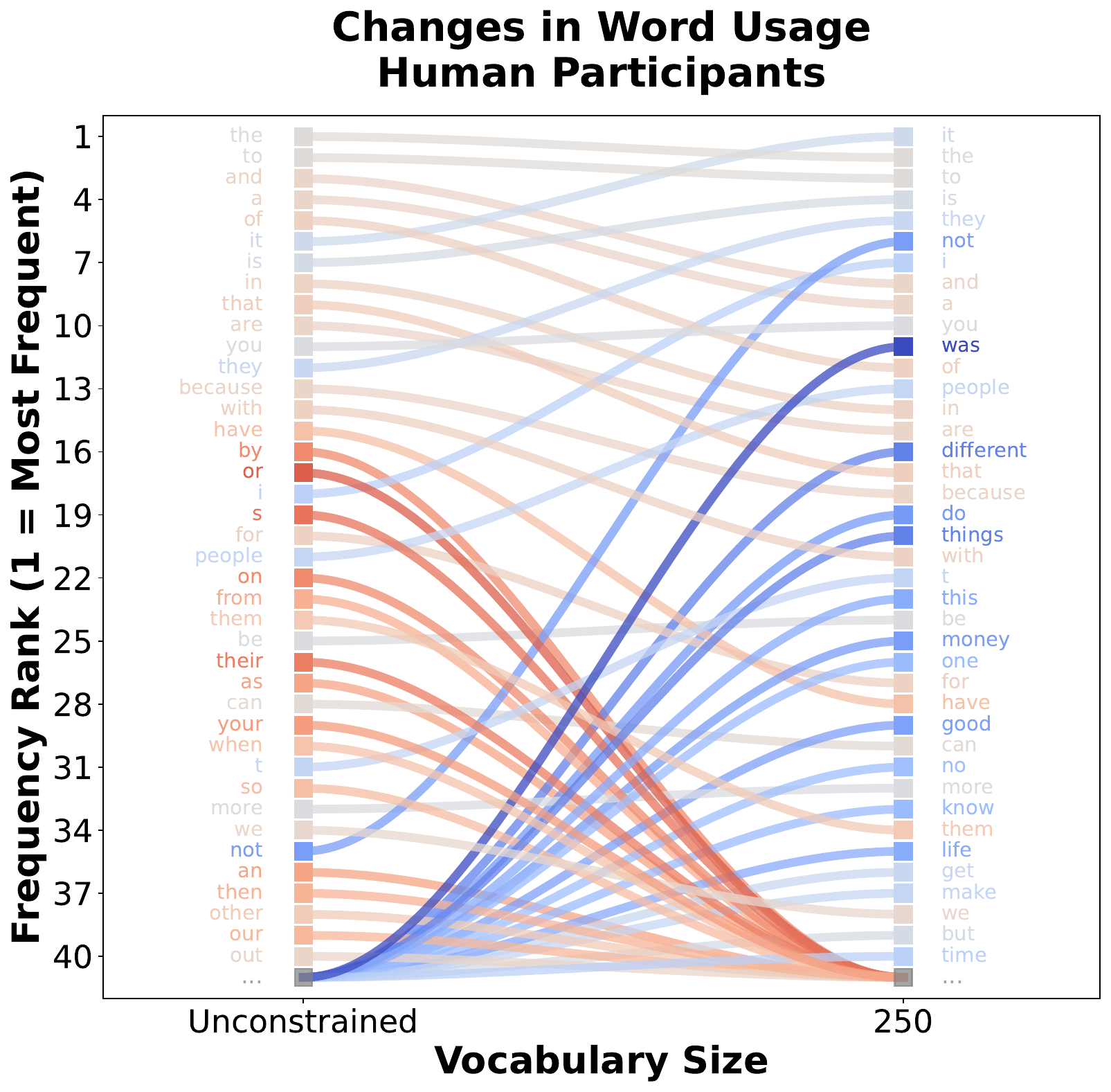}
  \caption {Bump plots for both models and humans showing word frequency rankings at largest and smallest vocabulary sizes. The color scale denotes sign and magnitude of rank change. Semantically light nouns, verbs, and adjectives (e.g., \textit{things}, \textit{make}, etc.) tend to rise in rank as vocabulary size shrinks, while connectives (\textit{or}, \textit{when}) tend to decrease in rank.}
  \label{fig:bump-plot}
\end{figure}

Consistent with our prediction, tighter vocabulary constraints induce responses with more frequent use of semantically light content words, such as \textit{thing, do,} or \textit{people}. Both human-generated and model-generated responses used semantically light words more frequently as the vocabulary constraint became stricter (\Cref{fig:bump-plot}). In particular, semantically light content words tended to rise in frequency rank, while function words tended to remain constant or fall in rank. We note that this qualitative pattern occurs for all model results, including \textsc{Awrs-16} and \textsc{Awrs-32}.

\subsection{Skilled humans revise more under constraint}

\begin{figure}[!htb]
\centering
  \includegraphics[width=0.9\columnwidth]{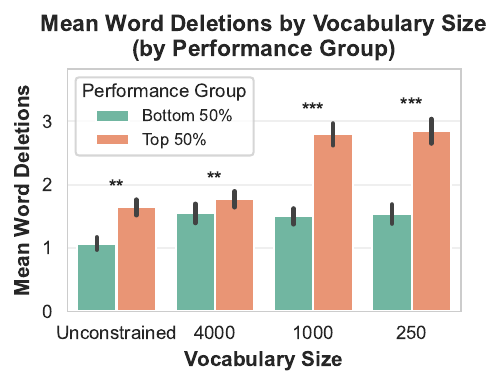}
  \caption {Mean valid word deletions per response, by vocabulary size and post-hoc performance group. Error bars denote 95\% confidence intervals. Stars denote p-values of Mann-Whitney U-test with Bonferroni correction (*: $<$0.05, **: $<$0.01, ***: $<$0.001).}
  \label{fig:word-deletions}
\end{figure}

\Cref{fig:word-deletions} shows the rate of the human respondents' word deletions per response by vocabulary size and participant skill group (a post-hoc split of the participants into the top and bottom 50\% by average score). For participants in the bottom 50\%, word deletions remained flat across different constrained vocabulary sizes. In contrast, for participants in the top 50\%, word deletions were significantly higher in the constrained vocabularies of 1000 and 250 words than in larger vocabularies. 
This much-greater propensity for revision among high-scoring humans suggests an attempt at a non-greedy approach, though these attempts may not be successful, as evidenced by the fact that the responses generated by this group of participants generally scored similarly to the greedy algorithm. 

\section{Discussion}

\subsection{Resource-rational inference in language production}

An extensive body of work in cognitive science and behavioral economics suggests that humans are approximately rational subject to cognitive costs \citep{kahnemanProspectTheoryAnalysis1979, kahnemanMapsBoundedRationality2003, griffithsRationalUseCognitive2015,liederResourcerationalAnalysisUnderstanding2020}.
In real-time communication, a pressure for communicative utility interacts with constraints on the availability of cognitive resources like time and attention.
Evidence also suggests that humans can adapt to \textit{novel} constraints by deploying strategies that make efficient use of available resources: people asked to communicate about novel objects rapidly converge to a set of labels, using shorter phrases for more frequently observed items \citep{kraussChangesReferencePhrases1964}. 
Studies have also shown emergent systematicity in highly constrained communication channels when there was a communicative goal, e.g. whistling to signal a color on a continuum \citep{chenDiscreteSystematicCommunication2025}. 

The intersection of language modeling and probabilistic inference provides an opportunity to compare human behavior to algorithmic-level hypotheses about processing. Sequential Monte Carlo has been argued to provide a cognitively plausible algorithm for approximate Bayesian inference for language comprehension \citep{clarkModelApproximateIncremental2025, clarkResourceRationalNoisyChannelLanguage2025}. 
The \textsc{AWRS} algorithm enables efficient \textit{generation} under a vocabulary constraint.
This algorithm is inherently resource-rational: it uses more samples when the constraint is difficult to satisfy than when it is easy, and varying the particle count results in a continuum of behavior from fast, greedy sampling to slower, more exact inference, which here resulted in higher-scoring responses.

Our results indicate that humans presented with the task of constrained generation show behavior that, somewhat surprisingly, more closely resembles greedy sampling than particle-based SMC. 
Above-average participants, however, show a signature of non-greedy sampling: they are much more likely to revise partially generated sentences. 
To help explain these results, we note that greedy sampling is one endpoint of a resource-rational trade-off; it minimizes computational effort but maximizes local bias. 
A plausible hypothesis is that humans are closer to the greedy extreme of this trade-off (possibly due to the high costs of revising or planning ahead), but that highly-motivated individuals may opt to increase effort to counter the dead ends induced by greedy sampling. 
Future work can consider whether financial incentives, in-person experimentation, or pairing participants together can induce more non-greedy behavior by placing greater weight on communicative success. 

Our results also indicate that certain words are used disproportionately when vocabulary is artificially restricted. In the most restricted vocabularies, semantically light content words jump substantially in usage rank. 
These words have especially high communicative utility in constrained settings, because they can serve as substitutes for low-frequency words (consider the frequent use of the word ``way'' in \Cref{tab:example_responses_8850}).
While intended words may not have perfect substitutes, they can often be replaced by a longer phrase, using relative clauses or prepositional phrases to add detail.
This pattern is also seen in both greedy and non-greedy model results, suggesting that it is a basic corollary of the substitutability of these words, rather than a result of active Bayesian inference.
These results are consistent with existing theoretical frameworks of built-in redundancy in language providing robustness against failures of communication \citep{tourtouriRationalRedundancyReferring2021, degenWhenRedundancyUseful2020,mahowaldGrammaticalCuesSubjecthood2023,leufkensFunctionalistTypologyRedundancy2020}.  

Another property of constrained generation with SMC is the sensitivity of inference quality and efficiency to the proposal distribution used to generate incremental next-word candidates. 
The greater the divergence between the proposal distribution and the target distribution, the more samples are needed to generate a word that meets the constraint; the algorithm works best when it does not have to reject too many next-word candidates \citep{lipkinFastControlledGeneration2025}. 
We speculate that this phenomenon has a natural analog in human cognition. There are two ways that a human can be good at constrained generation. The first is to take many possible ``samples''; we expect to see this reflected in slow processing and frequent deletions. The second is to have a good ``proposal distribution'' that assigns high probability to continuations which already satisfy the constraints; we expect this to be reflected in fluent production with relatively few revisions. 
Crucially, such a proposal distribution differs from the baseline distribution of language, and thus requires learning via experience. As people gain more familiarity with this task, it may become more natural and require less explicit inference, in keeping with theories of amortized computation in cognition \citep{gershmanAmortizedInferenceProbabilistic2014}. Future work may consider whether humans with extensive L2 experience are better than monolinguals at this task. 


\subsection{Intrinsic differences in question difficulty}

Are some questions simply harder to answer than others?
We list the top and bottom questions by average evaluator score for \textsc{Awrs-32} and for humans, when the vocabulary size was 250:

\begin{examplebox}{Top/Bottom AWRS-32 Questions}
\up\ Explain Christianity in simple terms. \\
\up\ Explain machine learning in simple terms. \\ 
\up\ Explain consequentialism in simple terms. \\
\up\ Explain the placebo effect in simple terms. \\
\down\ How do I make schnitzel? \\
\down\ Why does it echo if we yell in a cave but not a regular room? \\
\down\ How do you remove salt from water? \\
\down\ How do you prevent a sunburn?
\end{examplebox}

\begin{examplebox}{Top/Bottom Human Questions}
\up\ Explain the placebo effect in simple terms. \\
\up\ How do languages get new words? \\
\up\ How do maggots get into a place like a freezer that's sealed air tight when it loses power? \\
\up\ Why are human babies born so helpless? \\
\down\ How are books printed? \\
\down\ How can I haggle prices at a market effectively? \\
\down\ Explain the International Space Station in simple terms. \\
\down\ What is a hedge fund?
\end{examplebox}

First, we observe that for both the model and humans, most of the lowest-rated questions were in the How category.
This may be because How questions typically prompt mechanistic explanations, which are known to be hard to understand, even if the subject of the question is well-known or technically simple \citep[e.g.][]{mccarthyRightWayExplain2023, kelemenProfessionalPhysicalScientists2013, lombrozoFunctionalExplanationFunction2006, lombrozoMechanisticFunctionalUnderstanding2019}.
Additionally, these questions might involve lexicalized concepts that are difficult to explain via circumlocution.
For example, explaining how to make schnitzel is difficult without the word \textit{pork}, which is not among the most common 250 English words.
In contrast, the highest-rated questions likely present a communicator with alternative ways of expressing the idea, even if the question appears to be quite technical on the surface.
For example, the concept of the placebo effect can be expressed by words such as \textit{feel} and \textit{work}, both of which are among the 250 most frequent words.
The highest-scoring questions for \textsc{Awrs-32} all belong to the ExplainSimple dataset; this is possibly because the questions explicitly prompt the model to explain ``in simple terms'', which reduces the divergence between the proposal distribution and target constrained distribution.
These results suggest that whether a communicative goal is easy or hard to achieve under vocabulary constraints likely depends greatly on the availability of substitute words and well-suited analogies, which varies widely even within a question category, and not necessarily on the ``technical'' difficulty of the subject matter.

\subsection{Broader implications}

One key application of our work is for language learning. Our finding that semantically light words appear more frequently when the allowed vocabulary is small could provide useful guiding principles for optimizing language learning. For example, our results suggest that a foreign-language learner who expects to learn only 250 -- 1000 words in the target language should prioritize learning general-purpose, semantically light words (not necessarily the most frequent words) in order to be able to communicate effectively across a wide variety of everyday situations. 
In contrast, an intermediate-level speaker who hopes to sound more ``native'' might consider practicing communication without over-using semantically light words such as \textit{thing, do,} or \textit{people}, which occur less frequently in the unconstrained distribution of language.

Our results on the diminishing marginal returns of expanding one's vocabulary size also suggest that learning the first few hundred or thousand most frequent English words has disproportionately high returns for communicative utility (especially in the presence of a charitable comprehender), despite still being far below the vocabulary size of a native speaker. This could potentially explain why L2 learners often converge to an efficient but limited ``Basic Variety'' of a language \citep{kleinBasicVarietyCouldnt1997} or over-use familiar words \citep{hasselgrenLexicalTeddyBears1994}: if one can  communicate well using a restricted vocabulary, then there are fewer pressures to expand one's vocabulary. 


Additionally, modeling vocabulary-constrained communication has relevance for the study of language disorders. 
For example, communication for individuals with aphasia may benefit from flexible inference both in language production (choosing which words to produce when production is very difficult) and comprehension (how an interlocutor infers an intended meaning from fragmented utterances) \citep{beeke13AphasiaPragmatics2012, fedorenkoAgrammaticOutputNonfluent2022}.  
Patients with aphasia may be less likely to use low-frequency words like \textit{sailboat} while being more likely to use syntactically rich fragments like \textit{boat that is moved by the wind} to circumvent this limitation \citep{rezaiiSyntaxLexiconTradeoff2022}. 
Modeling how people generate language under constraint can shed light on the rapid, online inferences made by conversational partners in the presence of language disorders, and can potentially inform strategies and technologies for easing communication with vulnerable groups (for example, by modifying existing language models to better infer intended messages from atypical inputs).

This study has investigated how vocabulary constraints affect human-generated language, with theoretically-motivated comparisons to greedy and non-greedy sampling from language models using state-of-the-art algorithms. Our results demonstrate both the flexibility and the limitations of human language generated under vocabulary constraints, and provide an approach for computationally characterizing its properties. 


\bibliographystyle{apacite}

\setlength{\bibleftmargin}{.125in}
\setlength{\bibindent}{-\bibleftmargin}

\bibliography{CogSci_Template}

@incollection{beeke13AphasiaPragmatics2012,
  title = {13. {{Aphasia}}: {{The}} Pragmatics of Everyday Conversation},
  shorttitle = {13. {{Aphasia}}},
  booktitle = {13. {{Aphasia}}: {{The}} Pragmatics of Everyday Conversation},
  author = {Beeke, Suzanne},
  year = 2012,
  pages = {345--372},
  publisher = {De Gruyter Mouton},
  doi = {10.1515/9783110214215.345}
}

@article{bleakleyLanguageSkillsEarnings2004,
  title = {Language {{Skills}} and {{Earnings}}: {{Evidence}} from {{Childhood Immigrants}}*},
  shorttitle = {Language {{Skills}} and {{Earnings}}},
  author = {Bleakley, Hoyt and Chin, Aimee},
  year = 2004,
  journal = {The Review of Economics and Statistics},
  volume = {86},
  number = {2},
  pages = {481--496},
  issn = {0034-6535},
  doi = {10.1162/003465304323031067}
}

@incollection{bockLanguageProductionGrammatical1994,
  title = {Language Production: {{Grammatical}} Encoding},
  shorttitle = {Language Production},
  booktitle = {Handbook of Psycholinguistics},
  author = {Bock, Kathryn and Levelt, Willem},
  year = 1994,
  pages = {945--984},
  publisher = {Academic Press},
  address = {San Diego, CA, US}
}

@article{brysbaertHowManyWords2016,
  title = {How {{Many Words Do We Know}}? {{Practical Estimates}} of {{Vocabulary Size Dependent}} on {{Word Definition}}, the {{Degree}} of {{Language Input}} and the {{Participant}}'s {{Age}}},
  shorttitle = {How {{Many Words Do We Know}}?},
  author = {Brysbaert, Marc and Stevens, Micha{\"e}l and Mandera, Pawe{\l} and Keuleers, Emmanuel},
  year = 2016,
  journal = {Frontiers in Psychology},
  volume = {7},
  pages = {1116},
  issn = {1664-1078},
  doi = {10.3389/fpsyg.2016.01116},
  pmcid = {PMC4965448},
  pmid = {27524974}
}

@inproceedings{chandraCooperativeExplanationRational2024,
  title = {Cooperative Explanation as Rational Communication},
  booktitle = {Proceedings of the Annual Meeting of the Cognitive Science Society},
  author = {Chandra, Kartik and Chen, Tony and Li, Tzu-Mao and {Ragan-Kelley}, Jonathan and Tenenbaum, Josh},
  year = 2024,
  volume = {46}
}

@article{chenDiscreteSystematicCommunication2025,
  title = {Discrete and Systematic Communication in a Continuous Signal-Meaning Space},
  author = {Chen, Alicia M and Hofer, Matthias and Poliak, Moshe and Levy, Roger and Zaslavsky, Noga},
  year = 2025,
  journal = {Journal of Language Evolution},
  volume = {10},
  number = {1},
  pages = {lzaf003},
  issn = {2058-458X},
  doi = {10.1093/jole/lzaf003}
}

@article{clarkModelApproximateIncremental2025,
  title = {A {{Model}} of {{Approximate}} and {{Incremental Noisy-Channel Language Processing}}},
  author = {Clark, Thomas Hikaru and Vigly, Jacob Hoover and Gibson, Edward and Levy, Roger},
  year = 2025,
  journal = {Proceedings of the Annual Meeting of the Cognitive Science Society},
  volume = {47},
  number = {0}
}

@inproceedings{clarkResourceRationalNoisyChannelLanguage2025,
  title = {Resource-{{Rational Noisy-Channel Language Processing}}: {{Testing}} the {{Effect}} of {{Algorithmic Constraints}} on {{Inferences}}},
  shorttitle = {Resource-{{Rational Noisy-Channel Language Processing}}},
  booktitle = {Proceedings of the 2025 {{Conference}} on {{Empirical Methods}} in {{Natural Language Processing}}},
  author = {Clark, Thomas Hikaru and Vigly, Jacob Hoover and Gibson, Edward and Levy, Roger P.},
  editor = {Christodoulopoulos, Christos and Chakraborty, Tanmoy and Rose, Carolyn and Peng, Violet},
  year = 2025,
  pages = {23659--23672},
  publisher = {Association for Computational Linguistics},
  address = {Suzhou, China}
}

@article{cruzHowLaypeopleEvaluate2025,
  title = {How Laypeople Evaluate Scientific Explanations Containing Jargon},
  author = {Cruz, Francisco and Lombrozo, Tania},
  year = 2025,
  journal = {Nature Human Behaviour},
  volume = {9},
  number = {10},
  pages = {2038--2053},
  publisher = {Nature Publishing Group},
  issn = {2397-3374},
  doi = {10.1038/s41562-025-02227-0},
  copyright = {2025 The Author(s), under exclusive licence to Springer Nature Limited}
}

@article{degenWhenRedundancyUseful2020,
  title = {When Redundancy Is Useful: {{A Bayesian}} Approach to "Overinformative" Referring Expressions},
  shorttitle = {When Redundancy Is Useful},
  author = {Degen, Judith and Hawkins, Robert D. and Graf, Caroline and Kreiss, Elisa and Goodman, Noah D.},
  year = 2020,
  journal = {Psychological Review},
  volume = {127},
  number = {4},
  pages = {591--621},
  issn = {1939-1471},
  doi = {10.1037/rev0000186},
  pmid = {32237876}
}

@article{dornyeiCommunicationStrategiesSecond1997,
  title = {Communication {{Strategies}} in a {{Second Language}}: {{Definitions}} and {{Taxonomies}}},
  shorttitle = {Communication {{Strategies}} in a {{Second Language}}},
  author = {D{\"o}rnyei, Zolt{\'a}n and Scott, Mary Lee},
  year = 1997,
  journal = {Language Learning},
  volume = {47},
  number = {1},
  pages = {173--210},
  issn = {1467-9922},
  doi = {10.1111/0023-8333.51997005},
  copyright = {\copyright{} 1997 Language Learning Research Club, University of Michigan}
}

@article{fedorenkoAgrammaticOutputNonfluent2022,
  title = {Agrammatic Output in Non-Fluent, Including {{Broca}}'s, Aphasia as a Rational Behavior},
  author = {Fedorenko, Evelina and Ryskin, Rachel and Gibson, Edward},
  year = 2022,
  journal = {Aphasiology},
  volume = {0},
  number = {0},
  pages = {1--20},
  publisher = {Routledge},
  issn = {0268-7038},
  doi = {10.1080/02687038.2022.2143233}
}

@article{fedorenkoLanguagePrimarilyTool2024,
  title = {Language Is Primarily a Tool for Communication Rather than Thought},
  author = {Fedorenko, Evelina and Piantadosi, Steven T. and Gibson, Edward A. F.},
  year = 2024,
  journal = {Nature},
  volume = {630},
  number = {8017},
  pages = {575--586},
  publisher = {Nature Publishing Group},
  issn = {1476-4687},
  doi = {10.1038/s41586-024-07522-w},
  copyright = {2024 Springer Nature Limited}
}

@article{futrellInformationtheoreticPrinciplesIncremental2023,
  title = {Information-Theoretic Principles in Incremental Language Production},
  author = {Futrell, Richard},
  year = 2023,
  journal = {Proceedings of the National Academy of Sciences},
  volume = {120},
  number = {39},
  pages = {e2220593120},
  publisher = {Proceedings of the National Academy of Sciences},
  doi = {10.1073/pnas.2220593120}
}

@article{gershmanAmortizedInferenceProbabilistic2014,
  title = {Amortized {{Inference}} in {{Probabilistic Reasoning}}},
  author = {Gershman, Samuel and Goodman, Noah},
  year = 2014,
  journal = {Proceedings of the Annual Meeting of the Cognitive Science Society},
  volume = {36},
  number = {36}
}

@article{gibsonHowEfficiencyShapes2019,
  title = {How {{Efficiency Shapes Human Language}}},
  author = {Gibson, Edward and Futrell, Richard and Piantadosi, Steven P. and Dautriche, Isabelle and Mahowald, Kyle and Bergen, Leon and Levy, Roger},
  year = 2019,
  journal = {Trends in Cognitive Sciences},
  volume = {23},
  number = {5},
  pages = {389--407},
  issn = {1879-307X},
  doi = {10.1016/j.tics.2019.02.003},
  pmid = {31006626}
}

@article{griffithsRationalUseCognitive2015,
  title = {Rational {{Use}} of {{Cognitive Resources}}: {{Levels}} of {{Analysis Between}} the {{Computational}} and the {{Algorithmic}}},
  shorttitle = {Rational {{Use}} of {{Cognitive Resources}}},
  author = {Griffiths, Thomas L. and Lieder, Falk and Goodman, Noah D.},
  year = 2015,
  journal = {Topics in Cognitive Science},
  volume = {7},
  number = {2},
  pages = {217--229},
  issn = {1756-8757, 1756-8765},
  doi = {10.1111/tops.12142}
}

@misc{groggerWagePenaltyRegional2020,
  type = {Working {{Paper}}},
  title = {The {{Wage Penalty}} of {{Regional Accents}}},
  author = {Grogger, Jeffrey and Steinmayr, Andreas and Winter, Joachim},
  year = 2020,
  series = {Working {{Paper Series}}},
  number = {26719},
  eprint = {26719},
  publisher = {National Bureau of Economic Research},
  doi = {10.3386/w26719},
  archiveprefix = {National Bureau of Economic Research}
}

@misc{guSurveyLLMasaJudge2025,
  title = {A {{Survey}} on {{LLM-as-a-Judge}}},
  author = {Gu, Jiawei and Jiang, Xuhui and Shi, Zhichao and Tan, Hexiang and Zhai, Xuehao and Xu, Chengjin and Li, Wei and Shen, Yinghan and Ma, Shengjie and Liu, Honghao and Wang, Saizhuo and Zhang, Kun and Wang, Yuanzhuo and Gao, Wen and Ni, Lionel and Guo, Jian},
  year = 2025,
  number = {arXiv:2411.15594},
  eprint = {2411.15594},
  primaryclass = {cs},
  publisher = {arXiv},
  doi = {10.48550/arXiv.2411.15594},
  archiveprefix = {arXiv}
}

@article{hasselgrenLexicalTeddyBears1994,
  title = {Lexical Teddy Bears and Advanced Learners: A Study into the Ways {{Norwegian}} Students Cope with {{English}} Vocabulary},
  shorttitle = {Lexical Teddy Bears and Advanced Learners},
  author = {Hasselgren, Angela},
  year = 1994,
  journal = {International Journal of Applied Linguistics},
  volume = {4},
  number = {2},
  pages = {237--258},
  issn = {1473-4192},
  doi = {10.1111/j.1473-4192.1994.tb00065.x}
}

@article{honnibalSpaCyIndustrialstrengthNatural2020,
  title = {{{spaCy}}: {{Industrial-strength}} Natural Language Processing in {{Python}}},
  author = {Honnibal, Matthew and Montani, Ines and Van Landeghem, Sofie and Boyd, Adriane},
  year = 2020,
  doi = {10.5281/zenodo.1212303}
}

@article{kahnemanMapsBoundedRationality2003,
  title = {Maps of {{Bounded Rationality}}: {{Psychology}} for {{Behavioral Economics}}},
  shorttitle = {Maps of {{Bounded Rationality}}},
  author = {Kahneman, Daniel},
  year = 2003,
  journal = {American Economic Review},
  volume = {93},
  number = {5},
  pages = {1449--1475},
  issn = {0002-8282},
  doi = {10.1257/000282803322655392}
}

@article{kahnemanProspectTheoryAnalysis1979,
  title = {Prospect {{Theory}}: {{An Analysis}} of {{Decision}} under {{Risk}}},
  shorttitle = {Prospect {{Theory}}},
  author = {Kahneman, Daniel and Tversky, Amos},
  year = 1979,
  journal = {Econometrica},
  volume = {47},
  number = {2},
  eprint = {1914185},
  eprinttype = {jstor},
  pages = {263--291},
  publisher = {[Wiley, Econometric Society]},
  issn = {0012-9682},
  doi = {10.2307/1914185}
}

@article{keuleersWordKnowledgeCrowd2015,
  title = {Word Knowledge in the Crowd: {{Measuring}} Vocabulary Size and Word Prevalence in a Massive Online Experiment},
  shorttitle = {Word Knowledge in the Crowd},
  author = {Keuleers, Emmanuel and Stevens, Micha{\"e}l and Mandera, Pawe{\l} and Brysbaert, Marc},
  year = 2015,
  journal = {Quarterly Journal of Experimental Psychology},
  volume = {68},
  number = {8},
  pages = {1665--1692},
  address = {2006},
  issn = {1747-0226},
  doi = {10.1080/17470218.2015.1022560},
  pmid = {25715025}
}

@article{kleinBasicVarietyCouldnt1997,
  title = {The {{Basic Variety}} (or: {{Couldn}}'t Natural Languages Be Much Simpler?)},
  shorttitle = {The {{Basic Variety}} (Or},
  author = {Klein, Wolfgang and Perdue, Clive},
  year = 1997,
  journal = {Second Language Research},
  volume = {13},
  number = {4},
  pages = {301--347},
  publisher = {SAGE Publications Ltd},
  issn = {0267-6583},
  doi = {10.1191/026765897666879396}
}

@article{kraussChangesReferencePhrases1964,
  title = {Changes in Reference Phrases as a Function of Frequency of Usage in Social Interaction: A Preliminary Study},
  shorttitle = {Changes in Reference Phrases as a Function of Frequency of Usage in Social Interaction},
  author = {Krauss, Robert M. and Weinheimer, Sidney},
  year = 1964,
  journal = {Psychonomic Science},
  volume = {1},
  number = {1},
  pages = {113--114},
  issn = {2197-9952},
  doi = {10.3758/BF03342817}
}

@article{lauferDevelopmentL2Lexis1991,
  title = {The Development of {{L2}} Lexis in the Expression of the Advanced Learner},
  author = {Laufer, Batia},
  year = 1991,
  journal = {Modern Language Journal},
  volume = {75},
  number = {4},
  pages = {440--448},
  publisher = {Blackwell Publishing},
  address = {United Kingdom},
  issn = {1540-4781},
  doi = {10.2307/329493}
}

@article{leufkensFunctionalistTypologyRedundancy2020,
  title = {A Functionalist Typology of Redundancy},
  author = {Leufkens, Sterre},
  year = 2020,
  journal = {Revista da ABRALIN},
  pages = {79--103},
  issn = {0102-7158},
  doi = {10.25189/rabralin.v19i3.1722},
  copyright = {Copyright (c) 2020}
}

@misc{lewSequentialMonteCarlo2023,
  title = {Sequential {{Monte Carlo Steering}} of {{Large Language Models}} Using {{Probabilistic Programs}}},
  author = {Lew, Alexander K. and {Zhi-Xuan}, Tan and Grand, Gabriel and Mansinghka, Vikash K.},
  year = 2023,
  number = {arXiv:2306.03081},
  eprint = {2306.03081},
  primaryclass = {cs},
  publisher = {arXiv},
  doi = {10.48550/arXiv.2306.03081},
  archiveprefix = {arXiv}
}

@article{liederResourcerationalAnalysisUnderstanding2020,
  title = {Resource-Rational Analysis: {{Understanding}} Human Cognition as the Optimal Use of Limited Computational Resources},
  shorttitle = {Resource-Rational Analysis},
  author = {Lieder, Falk and Griffiths, Thomas L.},
  year = 2020,
  journal = {Behavioral and Brain Sciences},
  volume = {43},
  pages = {e1},
  publisher = {Cambridge University Press},
  issn = {0140-525X, 1469-1825},
  doi = {10.1017/S0140525X1900061X}
}

@misc{lipkinFastControlledGeneration2025,
  title = {Fast {{Controlled Generation}} from {{Language Models}} with {{Adaptive Weighted Rejection Sampling}}},
  author = {Lipkin, Benjamin and LeBrun, Benjamin and Vigly, Jacob Hoover and Loula, Jo{\~a}o and MacIver, David R. and Du, Li and Eisner, Jason and Cotterell, Ryan and Mansinghka, Vikash and O'Donnell, Timothy J. and Lew, Alexander K. and Vieira, Tim},
  year = 2025,
  number = {arXiv:2504.05410},
  eprint = {2504.05410},
  primaryclass = {cs},
  publisher = {arXiv},
  doi = {10.48550/arXiv.2504.05410},
  archiveprefix = {arXiv}
}

@article{lombrozoFunctionalExplanationFunction2006,
  title = {Functional Explanation and the Function of Explanation},
  author = {Lombrozo, Tania and Carey, Susan},
  year = 2006,
  journal = {Cognition},
  volume = {99},
  number = {2},
  pages = {167--204},
  issn = {0010-0277},
  doi = {10.1016/j.cognition.2004.12.009}
}

@misc{loulaSyntacticSemanticControl2025,
  title = {Syntactic and {{Semantic Control}} of {{Large Language Models}} via {{Sequential Monte Carlo}}},
  author = {Loula, Jo{\~a}o and LeBrun, Benjamin and Du, Li and Lipkin, Ben and Pasti, Clemente and Grand, Gabriel and Liu, Tianyu and Emara, Yahya and Freedman, Marjorie and Eisner, Jason and Cotterell, Ryan and Mansinghka, Vikash and Lew, Alexander K. and Vieira, Tim and O'Donnell, Timothy J.},
  year = 2025,
  number = {arXiv:2504.13139},
  eprint = {2504.13139},
  primaryclass = {cs},
  publisher = {arXiv},
  doi = {10.48550/arXiv.2504.13139},
  archiveprefix = {arXiv}
}

@article{mahowaldGrammaticalCuesSubjecthood2023,
  title = {Grammatical Cues to Subjecthood Are Redundant in a Majority of Simple Clauses across Languages},
  author = {Mahowald, Kyle and Diachek, Evgeniia and Gibson, Edward and Fedorenko, Evelina and Futrell, Richard},
  year = 2023,
  journal = {Cognition},
  volume = {241},
  pages = {105543},
  issn = {0010-0277},
  doi = {10.1016/j.cognition.2023.105543}
}

@book{marrVisionComputationalInvestigation1982,
  title = {Vision: {{A Computational Investigation}} into the {{Human Representation}} and {{Processing}} of {{Visual Information}}},
  shorttitle = {Vision},
  author = {Marr, David},
  year = 1982,
  publisher = {{Henry Holt and Co., Inc.}}
}

@article{mccarthyRightWayExplain2023,
  title = {A Right Way to Explain? {{Function}}, Mechanism, and the Order of Explanations},
  shorttitle = {A Right Way to Explain?},
  author = {McCarthy, Amanda M. and Keil, Frank C.},
  year = 2023,
  journal = {Cognition},
  volume = {238},
  pages = {105494},
  issn = {0010-0277},
  doi = {10.1016/j.cognition.2023.105494}
}

@article{pickeringIntegratedTheoryLanguage2013,
  title = {An Integrated Theory of Language Production and Comprehension},
  author = {Pickering, Martin J. and Garrod, Simon},
  year = 2013,
  journal = {The Behavioral and Brain Sciences},
  volume = {36},
  number = {4},
  pages = {329--347},
  issn = {1469-1825},
  doi = {10.1017/S0140525X12001495},
  pmid = {23789620}
}

@incollection{poulisseTheoreticalAccountLexical2011,
  title = {A {{Theoretical Account}} of {{Lexical Communication Strategies}}},
  booktitle = {The {{Bilingual Lexicon}}},
  author = {Poulisse, Nanda},
  year = 2011,
  pages = {157--190},
  publisher = {John Benjamins Publishing Company},
  chapter = {The Bilingual Lexicon},
  copyright = {De Gruyter expressly reserves the right to use all content for commercial text and data mining within the meaning of Section 44b of the German Copyright Act.}
}

@article{rezaiiSyntaxLexiconTradeoff2022,
  title = {A Syntax--Lexicon Trade-off in Language Production},
  author = {Rezaii, Neguine and Mahowald, Kyle and Ryskin, Rachel and Dickerson, Bradford and Gibson, Edward},
  year = 2022,
  journal = {Proceedings of the National Academy of Sciences of the United States of America},
  volume = {119},
  number = {25},
  doi = {10.1073/pnas.2120203119},
  pmcid = {9231468},
  pmid = {35709321}
}

@article{shannonMathematicalTheoryCommunication1948,
  title = {A Mathematical Theory of Communication},
  author = {Shannon, Claude E},
  year = 1948,
  pages = {55}
}

@misc{speerRspeerWordfreqV302022a,
  title = {Rspeer/Wordfreq: V3.0},
  shorttitle = {Rspeer/Wordfreq},
  author = {Speer, Robyn},
  year = 2022,
  doi = {10.5281/zenodo.7199437},
  howpublished = {Zenodo}
}

@article{sulikExplanationsWild2023,
  title = {Explanations in the Wild},
  author = {Sulik, Justin and {van Paridon}, Jeroen and Lupyan, Gary},
  year = 2023,
  journal = {Cognition},
  volume = {237},
  pages = {105464},
  issn = {0010-0277},
  doi = {10.1016/j.cognition.2023.105464}
}

@article{tourtouriRationalRedundancyReferring2021,
  title = {Rational {{Redundancy}} in {{Referring Expressions}}: {{Evidence}} from {{Event-related Potentials}}},
  shorttitle = {Rational {{Redundancy}} in {{Referring Expressions}}},
  author = {Tourtouri, Elli N. and Delogu, Francesca and Crocker, Matthew W.},
  year = 2021,
  journal = {Cognitive Science},
  volume = {45},
  number = {12},
  pages = {e13071},
  issn = {1551-6709},
  doi = {10.1111/cogs.13071},
  copyright = {\copyright{} 2021 The Authors. Cognitive Science published by Wiley Periodicals LLC on behalf of Cognitive Science Society (CSS).}
}

@misc{zhengJudgingLLMasaJudgeMTBench2023,
  title = {Judging {{LLM-as-a-Judge}} with {{MT-Bench}} and {{Chatbot Arena}}},
  author = {Zheng, Lianmin and Chiang, Wei-Lin and Sheng, Ying and Zhuang, Siyuan and Wu, Zhanghao and Zhuang, Yonghao and Lin, Zi and Li, Zhuohan and Li, Dacheng and Xing, Eric P. and Zhang, Hao and Gonzalez, Joseph E. and Stoica, Ion},
  year = 2023,
  number = {arXiv:2306.05685},
  eprint = {2306.05685},
  primaryclass = {cs},
  publisher = {arXiv},
  doi = {10.48550/arXiv.2306.05685},
  archiveprefix = {arXiv}
}

@article{ferreiraHowIncrementalLanguage2002,
    title = {How {Incremental} {Is} {Language} {Production}? {Evidence} from the {Production} of {Utterances} {Requiring} the {Computation} of {Arithmetic} {Sums}},
    volume = {46},
    issn = {0749-596X},
    shorttitle = {How {Incremental} {Is} {Language} {Production}?},
    url = {https://www.sciencedirect.com/science/article/pii/S0749596X01927974},
    doi = {10.1006/jmla.2001.2797},
    number = {1},
    urldate = {2026-04-09},
    journal = {Journal of Memory and Language},
    author = {Ferreira, Fernanda and Swets, Benjamin},
    month = jan,
    year = {2002},
    pages = {57--84},
}

@incollection{brewerExplanationScientistsChildren2000,
    address = {Cambridge, MA, US},
    title = {Explanation in scientists and children},
    isbn = {978-0-262-11249-9},
    booktitle = {Explanation and cognition},
    publisher = {The MIT Press},
    author = {Brewer, William F. and Chinn, Clark A. and Samarapungavan, Ala},
    year = {2000},
    pages = {279--298},
}

@article{kelemenProfessionalPhysicalScientists2013,
    address = {US},
    title = {Professional physical scientists display tenacious teleological tendencies: {Purpose}-based reasoning as a cognitive default},
    volume = {142},
    issn = {1939-2222},
    shorttitle = {Professional physical scientists display tenacious teleological tendencies},
    doi = {10.1037/a0030399},
    number = {4},
    journal = {Journal of Experimental Psychology: General},
    publisher = {American Psychological Association},
    author = {Kelemen, Deborah and Rottman, Joshua and Seston, Rebecca},
    year = {2013},
    pages = {1074--1083},
}

@incollection{lombrozoMechanisticFunctionalUnderstanding2019,
    title = {Mechanistic versus functional understanding},
    language = {eng},
    booktitle = {Varieties of {Understanding}: {New} {Perspectives} from {Philosophy}, {Psychology}, and {Theology}},
    publisher = {New York, NY: Oxford University Press},
    author = {Lombrozo, Tania and Wilkenfeld, Daniel A.},
    editor = {Grimm, Stephen R.},
    year = {2019},
    note = {149826
Section: 11},
    pages = {209--229},
}

@book{varianIntermediateMicroeconomicsModern2010,
    address = {New York},
    edition = {8},
    title = {Intermediate microeconomics: a modern approach},
    isbn = {978-0-393-93424-3 0-393-93424-1 978-0-393-93533-2 0-393-93533-7},
    url = {http://www.worldcat.org/search?qt=worldcat_org_all&q=0393934241},
    publisher = {W.W. Norton \& Co.},
    author = {Varian, Hal R.},
    year = {2010},
    note = {varian10
tex.added-at: 2015-02-10T03:45:35.000+0100
tex.interhash: 1075d5839aaa70f12ee85cfd5239e1f5
tex.intrahash: 8c9a8e25755e2c46b53db3505767e471
tex.refid: 317920200
tex.timestamp: 2015-02-10T03:55:59.000+0100},
}

@article{bullockJargonBarrierEffective2019,
    title = {Jargon as a barrier to effective science communication: {Evidence} from metacognition},
    volume = {28},
    issn = {0963-6625},
    shorttitle = {Jargon as a barrier to effective science communication},
    url = {https://doi.org/10.1177/0963662519865687},
    doi = {10.1177/0963662519865687},
    language = {EN},
    number = {7},
    urldate = {2026-04-09},
    journal = {Public Understanding of Science},
    publisher = {SAGE Publications Ltd},
    author = {Bullock, Olivia M. and Colón Amill, Daniel and Shulman, Hillary C. and Dixon, Graham N.},
    month = oct,
    year = {2019},
    pages = {845--853},
}

\end{document}